\documentclass[10pt,twocolumn,letterpaper]{article}

\usepackage[accsupp]{axessibility}  

\usepackage{cvpr}              
\usepackage{amsmath} 
\usepackage{amssymb,mathtools}  
\DeclareMathOperator*{\argmin}{arg\hspace{0.5mm}min\hspace{1mm}}
\usepackage{breqn}
\usepackage{multirow}
\usepackage[table,xcdraw]{xcolor}
\usepackage{cancel} 
\usepackage{xfrac}

\usepackage{arydshln}
{\endarray\hspace*{-0.5\arraycolsep}\endBmatrix}
\newenvironment{bMatrix}[1]{%
\bmatrix\array{#1}\hspace*{-0.5\arraycolsep}}%
{\endarray\endbmatrix}

\usepackage{longtable}
\usepackage{multirow, array} 
\usepackage{xcolor}
\usepackage{float}

\definecolor{cvprblue}{rgb}{0.21,0.49,0.74}
\usepackage[pagebackref,breaklinks,colorlinks,citecolor=cvprblue]{hyperref}

\def\paperID{4} 
\def\confName{CVPR}
\def\confYear{2024}

\title{Camera Motion Estimation from RGB-D-Inertial Scene Flow}

\author{Samuel Cerezo, Javier Civera\\
I3A,
        Universidad de Zaragoza\\
{\tt\small \{samueladriancerezo,jcivera\}@unizar.es}
}

\begin{document}
\maketitle
\begin{abstract}
In this paper, we introduce a novel formulation for camera motion estimation that integrates RGB-D images and inertial data through scene flow. Our goal is to accurately estimate the camera motion in a rigid 3D environment, along with the state of the inertial measurement unit (IMU). Our proposed method offers the flexibility to operate as a multi-frame optimization or to marginalize older data, thus effectively utilizing past measurements.
To assess the performance of our method, we conducted evaluations using both synthetic data from the ICL-NUIM dataset and real data sequences from the OpenLORIS-Scene dataset. Our results show that the fusion of these two sensors enhances the accuracy of camera motion estimation when compared to using only visual data.

\end{abstract}    
\section{Introduction}
\label{sec:intro}

Autonomous navigation plays a key role in enabling robots and various other applications, including mixed reality and autonomous driving. For that, precise motion estimates derived from onboard sensors are essential. 
And, in this context, scene flow stands as one of the fundamental techniques for motion estimation using RGB-D or range sensors ~\cite{jaimez2015fast,jaimez2018robust,zhang2020flowfusion,teed2021droid,rosinol2023nerf}.
More specifically, scene flow refers to the estimation of the 3D motion field of scene points obtained from two sensor readings~\cite{vedula_three-dimensional_1999}.
Although optical and scene flow have been used in numerous tasks over the years, such as motion compensation~\cite{zihao2018unsupervised}, object tracking~\cite{porzi2020learning} and object learning~\cite{prest2012learning}, we focus in this paper on the previously mentioned application of scene flow to odometry, \emph{i.e.}, the estimation of the camera motion.
We are also motivated by multisensor odometry and SLAM, which boost monocular-only approaches with extra accuracy and robustness, highly relevant in safety-critical robotic setups. Multisensor configurations have been widely explored in feature-based odometry and SLAM, e.g., stereo cameras \cite{pire2017s}, visual-inertial \cite{campos2021orb} or LiDAR-inertial \cite{shan2020lio}, but are less explored in direct approaches that use the raw sensor measurements without feature extraction. 

On the one hand, RGB-D cameras provide a practical hardware alternative to several challenges and limitations of visual odometry. Their availability at low cost has facilitated many robotics and Augmented Reality (AR) applications in the last decade. 
Today, RGB-D cameras stand out as one of the preferred sensors for indoor applications in robotics and AR; and their future looks promising either on their own or in combination with additional sensors.
On the other hand, most commercial mobile devices are equipped with Inertial Measurement Units (IMU), which can provide large amount of information in dynamic trajectories but exhibit large drift due to noises if not fused with other information. This makes the visual-range-inertial fusion relevant, as the three  modalities offer complementary characteristics. 


Our contribution in this paper is a RGB-D-inertial formulation for camera motion estimation in rigid scenes. Up to our knowledge, this is the first time that inertial data is fused together with color and depth measurements to estimate camera motion based on optical flow. Specifically, we propose a tightly coupled optimization by minimizing pre-integrated inertial residuals and range constraints. As the inertial states are common between frames, we formulate the problem as a multi-frame optimization, in which past frame's states can be estimated or marginalized out into prior residuals for the inertial states. We evaluate our proposal in the synthetic ICL-NUIM dataset and in the real OpenLORIS-Scene one. The effectiveness of our fusion is shown by an error reduction of RGB-D-inertial estimation compared to RGB-D one.

\section{Related Work}
\label{sec:related}




\indent The first tracking system for ego-motion estimation which fuses vision and inertial measurements was presented by Armesto \etal \cite{FastEgo-Estimation2007}. In this case, the fusion is performed by considering a EKF and UKF (Extended and Unscented Kalman Filters) with multi-rate sampling of measurements. The mentioned sampling modality allows the system to work with the different rates of the sensors. In 2013, Kerl \etal \cite{kerl_robust_2013} proposed a fast and accurate method to estimate the camera motion from RGB-D images. This approach estimates the relative motion between two consecutive frames by minimizing the photometric error. 
A motion prior is incorporated in the optimization, in order to guide and stabilize motion estimation in the presence of dynamic objects.
Nie{\ss}ner \etal \cite{niessner2014combining} developed an approach that improves the robustness of real-time 3D surface reconstruction by incorporating inertial sensing to the inter-frame alignment. As a result, they could significantly reduce the number of Iterative Closest Point (ICP) iterations required per frame.
Modeling three-dimensional scene motion as a twist field, Quiroga \etal \cite{quiroga_dense_2014} introduced a method that encourages piecewise smooth solutions of rigid body motions. A general formulation is given to solve local and global rigid motions by jointly using intensity and depth data.

The first method to compute dense scene flow in real-time for RGB-D cameras was introduced in 2015 by Jaimez \etal \cite{jaimez2015primal}. They proposed a variational formulation where brightness and geometric consistency are imposed. Their accuracy outperforms that of previous for RGB-D flow baselines, being able to estimate non-rigid motions at 30Hz of frame rate. In the same year, 
Leutenegger \etal \cite{leutenegger_keyframe-based_2015} formulated a probabilistic cost function that combines reprojection errors of landmarks with inertial terms, using stereo and monocular cameras. 
On the other hand, a new dense method to compute the odometry of a range sensor in real time is presented \cite{jaimez2015fast}. This method applies the range flow constraint equation in order to obtain the velocity of the sensor in a rigid environment. Experiments show that this approach overperforms GICP which uses the same geometric input data, whereas it achieves results similar to RDVO, which requires both geometric and photometric data.

The first tightly-coupled dense RGB-D-inertial SLAM system was proposed in 2017 by Laidlow \etal~\cite{laidlow_dense_2017}. This system jointly optimises the camera pose, velocity, IMU biases and gravity direction while building up a globally consistent, fully dense surfel-based 3D reconstruction of the environment.
In 2019, Shan \etal introduced VINS-RGBD \cite{shan_rgbd-inertial_2019}. The authors integrate a mapping system based on depth data and octree filtering to achieve real-time mapping. However the proposed system is applied only in ground robots.
A RGB-D scene flow estimation method with global nonrigid and local rigid motion assumption is proposed by Li \etal in \cite{li_rgbd_2020}. 
3D motion is estimated based on the global non-rigid and local rigid assumption and spatial-temporal correlation of RGBD information. 
With this assumption, the interaction of motion from different parts in the same segmented region is avoided, especially the non-rigid object, e.g., a human body. 

The flow formulation has been adapted to novel sensor modalities, e.g., event cameras~\cite{schnider2023neuromorphic}, or to include additional information, such as robot dynamics in the work of Lee \etal~\cite{lee2020aggressive}. Similarly to ours, the motivation in this last case is improving the robustness and accuracy of the camera motion estimation. Zhai \etal \cite{zhai2021optical} compiled in a survey recent advances on optical and scene flow. Up to our knowledge, inertial sensing has never been integrated in flow formulations. Our work contributes to the literature presenting the first camera motion estimation from RGB-D-inertial scene flow, demonstrating its effectiveness in simulated and real public datasets.

\section{Notation}
\label{sec:notation}



\indent Throughout this article, bold lower-case letters ($\mathbf{x}$) represent vectors and bold upper-case letters ($\boldsymbol{\Sigma}$) matrices. Scalars will be represented by light lower-case letters (${\alpha}$), scalar functions and images by light upper-case letters ($J$).
Camera poses are represented as $\mathbf{T}_{WB}=[\mathbf{R}_{WB} ,\hspace{0.5mm} ^W\mathbf{p}] \in {SE}(3)$ and transform points from frame $B$ to world coordinate system $W$. 

\section{IMU Model and Motion Integration}
\label{sec:IMU-integration}


\subsection{Inertial preintegration}

An IMU consists typically of an accelerometer and a three-axis gyroscope, and measures the angular velocity $^{B}\boldsymbol{\omega}$ and linear acceleration $^{B}\mathbf{a}$ of the sensor in the body reference frame $B$.
We will denote the IMU measurement at time $k$ as $^{B}\Tilde{\boldsymbol{\omega}}_{k}$ and $^{B}\Tilde{\mathbf{a}}_k$. IMU measurements are affected by additive white noise $\boldsymbol{\eta}^{g}$, $\boldsymbol{\eta}^{a} \in \mathbb{R}^{3}$ and two slowly varying gyroscope and accelerometer bias $\mathbf{b}^g$ and $\mathbf{b}^a \in \mathbb{R}^{3}$ respectively. Finally, the acceleration measurement is affected by gravity $^{W}\mathbf{g}$. This model is formulated by Eq. (\ref{eq:imu_model1}) and (\ref{eq:imu_model2}). 
\begin{equation}
\label{eq:imu_model1}
        ^{B}\Tilde{\boldsymbol{\omega}}_k =\hspace{0.5mm} ^{B}\boldsymbol{\omega}_k + \mathbf{b}^g_k + \boldsymbol{\eta}^{g}_k
\end{equation}
\begin{equation}
\label{eq:imu_model2}
        ^{B}\Tilde{\mathbf{a}}_k = \mathbf{R}_{WB}^{\top}\left(^{W}\mathbf{a}_k -\hspace{0.5mm} ^{W}\mathbf{g}\right) + \mathbf{b}^a_k + \boldsymbol{\eta}^{a}_k
\end{equation}
We use pre-integrated inertial residuals as proposed by Forster \emph{et al.} \cite{forster-manifold-2017}. We compute an inital guess for $\mathbf{b}^g$ as the difference of an estimate $^{B}\boldsymbol\omega_k$ of the angular velocity between two consecutive frames, and the direct measurement of the gyroscope $^{B}\Tilde{\boldsymbol\omega}_k$ that includes the bias, \emph{i.e.}, $\hat{\mathbf{b}}^g_k =\hspace{0.5mm}^{B}\Tilde{\boldsymbol\omega}_k -\hspace{0.5mm} ^{B}\boldsymbol\omega_k$, 
where the angular velocity estimate $^{B}\boldsymbol\omega_k$ can be computed by relative motion estimation between point clouds, divided by the time increment between them. We set the initial seed for $\mathbf{b}^a_k$ to zero.

Following \cite{forster-manifold-2017}, we can use the relative motion increment $\Delta \mathbf{v}_{ij} \doteq \hspace{0.5mm} \mathbf{R}_i^{\top}(\mathbf{v}_j - \mathbf{v}_i - \mathbf{g}\Delta t_{ij})$ in order to obtain a first gravity vector estimation 
\begin{equation}
\label{eq:gravity-vector-esstimation}
\hat{\mathbf{g}} = \frac{\mathbf{v}_j - \mathbf{v}_i}{\Delta t_{ij}} - \frac{\mathbf{R}_i\Delta \mathbf{v}_{ij}}{\Delta t_{ij}}  \end{equation}

\subsection{Noise propagation}
The covariance matrix of the raw IMU measurements noise $\boldsymbol{{\Sigma}_{\eta}} \in \mathbb{S}^{6}_+$\footnote{By $\mathbb{S}^{n}_+=\{\mathbf{\Sigma} \in \mathbb{R}^{n \times n} \ | \ \mathbf{\Sigma} = \mathbf{\Sigma}^\top, \mathbf{\Sigma}\succeq 0\}$ we denote the set of $n \times n$ symmetric positive semidefinite matrices.} is composed by sub-matrices
$\boldsymbol{{\Sigma}_{\omega}},\boldsymbol{{\Sigma}_{\textbf{a}}} \in \mathbb{S}^{3}_+$
\begin{equation}
\label{eq:matriz-covarianza-ruido}
 \boldsymbol{{\Sigma}_{\eta}}
  =\begin{bMatrix}{c:c}
  \boldsymbol{{\Sigma}_{\omega}} & \textbf{0}_{3\mathrm{x}3}\\ \hdashline \textbf{0}_{3\mathrm{x}3} & \boldsymbol{{\Sigma}_{\textbf{a}}}
  \end{bMatrix}
\end{equation}

\noindent where $\textbf{0}_{3\mathrm{x}3}$ is a $3\mathrm{x}3$ matrix for which all its elements are equal to zero.

Following the computation of the preintegrated noise covariance in \cite{forster-manifold-2017}, we consider the matrix $\boldsymbol\eta_{ik}^{\Delta} \doteq \left[\delta \boldsymbol\phi_{ik}^\top , \delta \mathbf{v}_{ik}^\top , \delta   \mathbf{p}_{ik}^\top\right]^\top$
and defining the IMU measurement noise $\boldsymbol\eta_{k}^{d} \doteq \left[\boldsymbol\eta_{k}^{gd} , \boldsymbol\eta_{k}^{ad}\right]^\top$, 
the noise is propagated as
\begin{equation}
\label{eq:def-preintegrated-measurement-covariance}
        {\boldsymbol\Sigma}_{ij} = \mathbf{A}_{j-1}{\boldsymbol\Sigma}_{ij-1}\mathbf{A}_{j-1}^\top +  \mathbf{B}_{j-1}\boldsymbol{{\Sigma}_{\eta}}\mathbf{B}_{j-1}^\top
\end{equation}
\noindent with initial conditions ${\boldsymbol\Sigma}_{ii} = \textbf{0}_{9\times 9}$ and $\mathbf{A}_{j-1} \in \mathbb{R}^{9\times 9}$ , $\mathbf{B}_{j-1} \in \mathbb{R}^{9\times6}$  defined as
\begin{equation*}
\mathbf{A}_{j-1}=\begin{bmatrix}
\Delta \Tilde{\mathbf{R}}_{j-1j}^\top & \textbf{0}_{3\mathrm{x}3} & \textbf{0}_{3\mathrm{x}3}\\
-\Delta \Tilde{\mathbf{R}}_{ij-1}(\Tilde{\mathbf{a}}_{j-1} - \mathbf{b}_i^a)^{\wedge}\Delta t & \textbf{I}_{3} & \textbf{0}_{3\mathrm{x}3}\\
-\frac{1}{2}\Delta \Tilde{\mathbf{R}}_{ij-1}(\Tilde{\mathbf{a}}_{j-1} - \mathbf{b}_i^a)^{\wedge}\Delta t^2  & \textbf{I}_{3}\Delta t & \textbf{I}_{3}
\end{bmatrix}
\end{equation*}
\begin{equation*}
\mathbf{B}_{j-1}=\begin{bmatrix}
\mathbf{J}_r^{j-1} \Delta t & \textbf{0}_{3\mathrm{x}3}\\
\textbf{0}_{3\mathrm{x}3} & \Delta \Tilde{\mathbf{R}}_{ij-1} \Delta t\\
\textbf{0}_{3\mathrm{x}3} & \frac{1}{2}\Delta \Tilde{\mathbf{R}}_{ij-1} \Delta t^2
\end{bmatrix}
\end{equation*}

\noindent where $\textbf{I}_{3}$ stands for the identity matrix of size $3$.

For the computation of the above matrices, the preintegrated expressions in \cite{forster-manifold-2017} were used.
The matrix ${\boldsymbol\Sigma}_{ij} \in \mathbb{S}^{9}_+$ is composed by nine sub-matrices of dimension 3 by 3 each
\begin{equation}
\label{eq:submatrices-matriz-Sigma}
 \boldsymbol{{\Sigma}}_{ij}
  =\begin{bMatrix}{c:c:c}
  {\boldsymbol\Sigma}_{\Delta \boldsymbol\phi_{ij}} & {\boldsymbol\Sigma}_{\Delta \boldsymbol\phi_{ij}\Delta \mathbf{v}_{i}} & {\boldsymbol\Sigma}_{\Delta \boldsymbol\phi_{ij}\Delta   \mathbf{p}_{ij}}\\ \hdashline {\boldsymbol\Sigma}_{\Delta \mathbf{v}_{i}\Delta \boldsymbol\phi_{ij}} & {\boldsymbol\Sigma}_{\Delta \mathbf{v}_{i}} & {\boldsymbol\Sigma}_{\Delta \mathbf{v}_{i}\Delta   \mathbf{p}_{ij}}\\ \hdashline {\boldsymbol\Sigma}_{\Delta   \mathbf{p}_{ij}\Delta \boldsymbol\phi_{ij}} & {\boldsymbol\Sigma}_{\Delta   \mathbf{p}_{ij}\Delta \mathbf{v}_{i}} & {\boldsymbol\Sigma}_{\Delta   \mathbf{p}_{ij}}
  \end{bMatrix}
\end{equation}

The above matrix will be important in the calculation of inertial residuals in the next section.

\subsection{Gravity vector representation}

As the gravity modulus is known, a reasonable representation is by its directional vector. Unit-norm direction vectors belong to the ${S}^{2}$ manifold, which has only two degrees of freedom. As ${S}^{2}$ does not form a Lie group, we follow the parametrization proposed in \cite{hertzberg2013integrating}.
The explicit expressions are detailed in \cite{teach-and-repeat}.


\subsection{Optical flow and velocity constraint}

We adopt the standard assumption that the local intensity image patterns are approximately constant, at least in the short period of time between two frames of a video~\cite{beauchemin_optical_computation_1995}. This constraints the motion in the image as in the following
\begin{equation}
\label{eq_optical_flow}
   0 = \frac{\partial I}{\partial t} + \frac{\partial I}{\partial u} {\dot u} + \frac{\partial I}{\partial v} {\dot v}
\end{equation}

\noindent where $(\dot u, \dot v)$ is the optical flow in image units ($\text{pixels}/s$).
Under the common assumption of a rigid scene, we formulate the point velocities in terms of the camera motion.
Let $Z: \Omega \rightarrow \mathbb{R}$ be a depth image provided by a 3D range camera where $\Omega$ is the image domain. Following the work by Spies \textit{et al.} \cite{SpiesJB02}, the range flow constraint is as follows
\begin{equation}
\label{eq_range_flow_constraint_demo}
    {\dot Z} = ({\dot w})=  \frac{\partial Z}{\partial t} +\frac{\partial Z}{\partial u} {\dot u} + \frac{\partial Z}{\partial v} {\dot v}
\end{equation}

This equation reflects that the total derivative of the depth can be calculated from the optical flow and the partial derivatives of $Z$. Following \cite{jaimez2015fast} and using the pin-hole model, 
we obtain the range flow constraint in Eq. (\ref{eq:velocity-constraint}). Here $f_x , f_y$ are the focal length values, expressed in pixels, while $\dot x, \dot y, \dot z$ are in camera coordinates.

\begin{eqnarray}
\label{eq:velocity-constraint}
        -\frac{\partial Z}{\partial t} &=& \left( 1 + \frac{x f_x}{z^{2}} \frac{\partial Z}{\partial u} + \frac{y f_y}{z^{2}} \frac{\partial Z}{\partial v} \right)\left(v_z + y \omega_x - x \omega_y\right) \nonumber \\ 
         & & + \frac{f_x}{z}\frac{\partial Z}{\partial u}(-v_x + y \omega_z - z \omega_y) \\
         & & + \frac{f_y}{z}\frac{\partial Z}{\partial v}(-v_y - x \omega_z + z \omega_x) \nonumber
\end{eqnarray}

The above constraint for the camera velocity will be used for our visual residuals, detailed in next section.

\section{Camera Motion from RGB-D-I Flow}
\label{sec:RGBD-inertial}

This section presents our approach to integrating inertial measurements with RGB-D scene flow to estimate camera motion. We begin by defining the state, which varies according to different operating modes, primarily depending on the number of frames considered. We then proceed to formulate the cost function to be optimized. Finally, this section concludes with the marginalization process, through which we retain the information of removed states.

\subsection{State definition}

\begin{figure}[H]
\centerline{\includegraphics[scale=0.22]{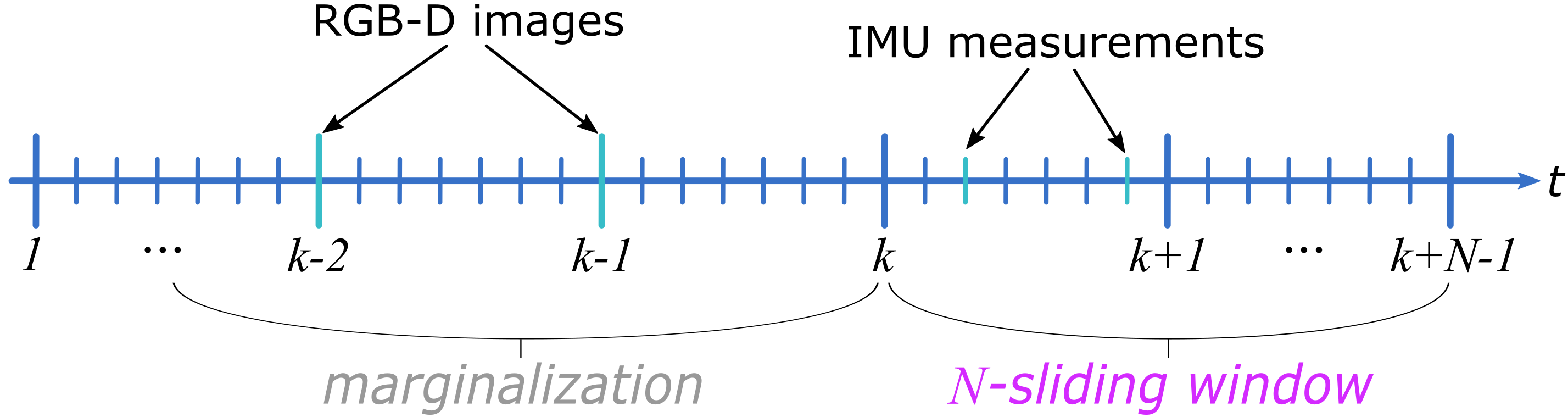}}
\caption{Illustration of the temporal notation for RGB-D images, IMU measurements and marginalization and optimization windows.}
\label{linea-temporal}
\end{figure}

Our goal is to track the state $\mathbf{x}$ of a sensing device equipped with an IMU and a RGB-D camera.
This state consists basically of the device velocities, IMU biases and gravity vector at different moments of time.
We assume that the IMU is synchronized with the camera, as it is shown in Fig. \ref{linea-temporal}.

In a general case, the system performs the optimization over \textit{N} frames, what we called a \textit{N-sliding window}.
As the sensing device moves through the trajectory, we marginalize out the old states for the optimization to remain compact. 
The state at step $k+N-1$ is defined as 
\begin{equation}
        \mathbf{x} = \left[{\mathbf{v}}_k^\top,{\boldsymbol\omega}_k^\top, \cdots ,{\mathbf{v}}_{k+N-1}^\top,{\boldsymbol\omega}_{k+N-1}^\top,\mathbf{g}^\top,{\mathbf{b}^g}^\top, {\mathbf{b}^a}^\top \right]^\top
\end{equation}
\noindent where $\{\mathbf{v}_l\}^{k+N-1}_{l=k} \in \mathbb{R}^{3}$ are the linear velocities in each frame, $\{\boldsymbol\omega_l\}^{k+N-1}_{l=k} \in \mathbb{R}^{3}$ are the angular velocities in each frame, $\mathbf{g} \in  \mathrm{S}^{2}$ contain the two degrees of freedom of the gravity direction and $\mathbf{b}^g, \mathbf{b}^a \in \mathbb{R}^{3}$ 
are the gyroscope and accelerometer bias, respectively.

As mentioned, we include the gravity direction as part of the state. Under the traditional absolute formulations for visual-inertial state estimation, this variable is removed from the state by aligning the global reference frame to the gravity direction during system initialization. But, as a consequence, the rest of the variables in the state are tied to a gravity-aligned absolute frame. In turn, by including the gravity vector in the local camera frames we remove this dependence and all states are relative. As an additional benefit, it becomes possible to explicitly re-estimate the gravity direction during normal system operation and thus avoid coupling gravity and absolute orientation errors. In order to improve the observability of the state (in particular of the accelerometer bias), we assume a known gravity magnitude (981 $cm/s^2$) and only optimize the gravity direction.

\subsection{Cost function}

We formulate an optimization problem over the state $\mathbf{x}$
for which the camera velocity consistency is imposed as well as those terms corresponding to the pre-integration of the IMU readings.
The joint optimization problem will consist on minimizing a cost function $J(\mathbf{x})$ which is the summation of terms associated to the inertial measurements $J_{i}$ as well as to the camera measurements $J_{c}$.
Our state estimate $\hat{\mathbf{x}}$ will be the one that minimizes the cost function $J(\mathbf{x})$.
\begin{align}
\label{eq:optimizacion_problem}
        \hat{\mathbf{x}} = \argmin_{\mathbf{x}}  J(\mathbf{x}) = \argmin_{\mathbf{x}}  \left(  J_{c}(\mathbf{x}) + J_{i}(\mathbf{x}) \right)
\end{align}

We begin by developing the term $J_{c}$. For a pair of consecutive frames $i$ and $j$, the velocity constraint in Eq. (\ref{eq:velocity-constraint}) results in the linear constraint 
\begin{equation}
\label{eq:primera-eq-res-visual}
        \mathbf{r}_{c} =   \mathbf{WAx} - \mathbf{WB}
\end{equation}

\noindent where $\mathbf{A}$ contains the weights of the coefficients that multiply the state vector $\mathbf{x}$ in the velocity constraint, and the matrix $\mathbf{B}$ contains the temporal derivatives of the per-pixel depths (inverted in sign). The linearization that is applied to derive the range flow constraint in Eq. (\ref{eq:velocity-constraint}) assumes differentiability of the depth images and small scene displacement. Therefore, we implement an adaptive mask on the image, in order to discard those pixels belonging to edges and prone to have high depth derivatives. 
This mask is represented in a diagonal matrix
$\mathbf{W}$, which also has the weights associated with the uncertainty of each equation. For details on these aspects, the reader is referred to \cite{jaimez2015fast}.
Using the residual in Eq. (\ref{eq:primera-eq-res-visual}), the visual cost is expressed as

\begin{equation}
\label{eq:expresion-visual-residual}
    J_{c} = \mathbf{r}_{c}^\top{\boldsymbol\Sigma}_{c}^{-1}\mathbf{r}_{c}
\end{equation}

Having developed the first term of the cost function, we now turn our attention to the development of $J_{i}$, which will be composed by several terms. Firstly, from Eq. (\ref{eq:imu_model1}) we can derive the residual associated to the angular velocity estimate, as follows
\begin{equation}
\label{eq:ang-velocity-estimation}
        \mathbf{r}_{\omega} =\hspace{0.5mm}  ^{B}\boldsymbol\omega_{} - \left(^{B}\Tilde{\boldsymbol\omega}_{} - \mathbf{b}^g\right)
\end{equation}

Assuming constant biases, as we compute flow for a small number of frames, the residual for the preintegrated linear velocity term is defined in \cite{forster-manifold-2017} as 
\begin{equation}
\label{eq:residual-delta-V-forster}
       \mathbf{r}_{\Delta \mathbf{v}_{i}} =\hspace{0.5mm} \mathbf{R}_i^{\top}(\mathbf{v}_j - \mathbf{v}_i - \mathbf{g}\Delta t_{}) - \Delta \Tilde{\mathbf{v}}_{ij}
\end{equation}

The residual for biases are made by penalising changes as the sliding window moves \emph{i.e.}, $\mathbf{r}_{{ba}} =  \mathbf{b}_0^a - \mathbf{b}^a$ and $\mathbf{r}_{{bg}} =  \mathbf{b}_0^g - \mathbf{b}^g$, where  $\mathbf{b}_0^a$ and $\mathbf{b}_0^g$ are the initial estimates.

Using the submatrices ${\boldsymbol\Sigma}_{\Delta \mathbf{v}_{i}}, {\boldsymbol\Sigma}_{\boldsymbol\omega}, {\boldsymbol\Sigma}_{\mathbf{a}} \in \mathbb{S}^{3}_+$ from Eq. (\ref{eq:submatrices-matriz-Sigma}), we can now define the term $J_i$ as follows
\begin{eqnarray}
\label{eq:expresion-inertial-residual}
J_{i} &=& 
\mathbf{r}_{\Delta \mathbf{v}_{i}}^\top {\boldsymbol\Sigma}_{\Delta \mathbf{v}_{i}}^{-1} \mathbf{r}_{\Delta \mathbf{v}_{i}} + \mathbf{r}_{\omega}^\top{\boldsymbol\Sigma}_{\boldsymbol\omega}^{-1}\mathbf{r}_{\omega} \nonumber \\
& & +\mathbf{r}_{{bg}}^\top{\boldsymbol\Sigma}_{\boldsymbol\omega}^{-1}\mathbf{r}_{{bg}}
+\mathbf{r}_{{ba}}^\top{\boldsymbol\Sigma}_{\mathbf{a}}^{-1} \mathbf{r}_{{ba}}
\end{eqnarray}

Up to this point we have defined the cost function, by means of $J_c$ and $J_i$.
In the general case $J(\mathbf{x})$ will be made up depending on the number of frames in each case. These cases will be detailed below.

\subsection{Operating Modes}
\begin{figure}[h]
\centerline{\includegraphics[scale=0.14]{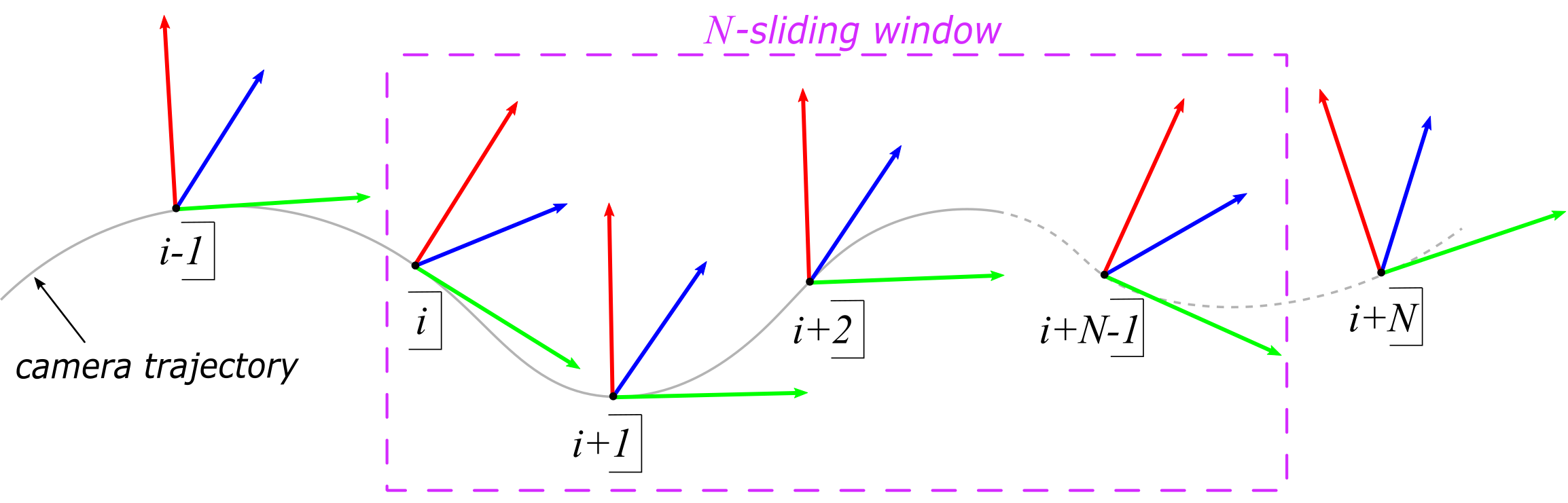}}
\caption{Optimization using a sliding window over \textit{N} frames while the camera is moving over the trajectory.}
\label{optimization_N_frames}
\end{figure}

We have mentioned that the cost function $J(\mathbf{x})$ will be formed as a function of the number of frames ($N$) in the sliding window while the camera is moving along the trajectory. Fig. \ref{optimization_N_frames} illustrates this situation graphically.

\begin{figure}[h]
\centerline{\includegraphics[scale=1.83]{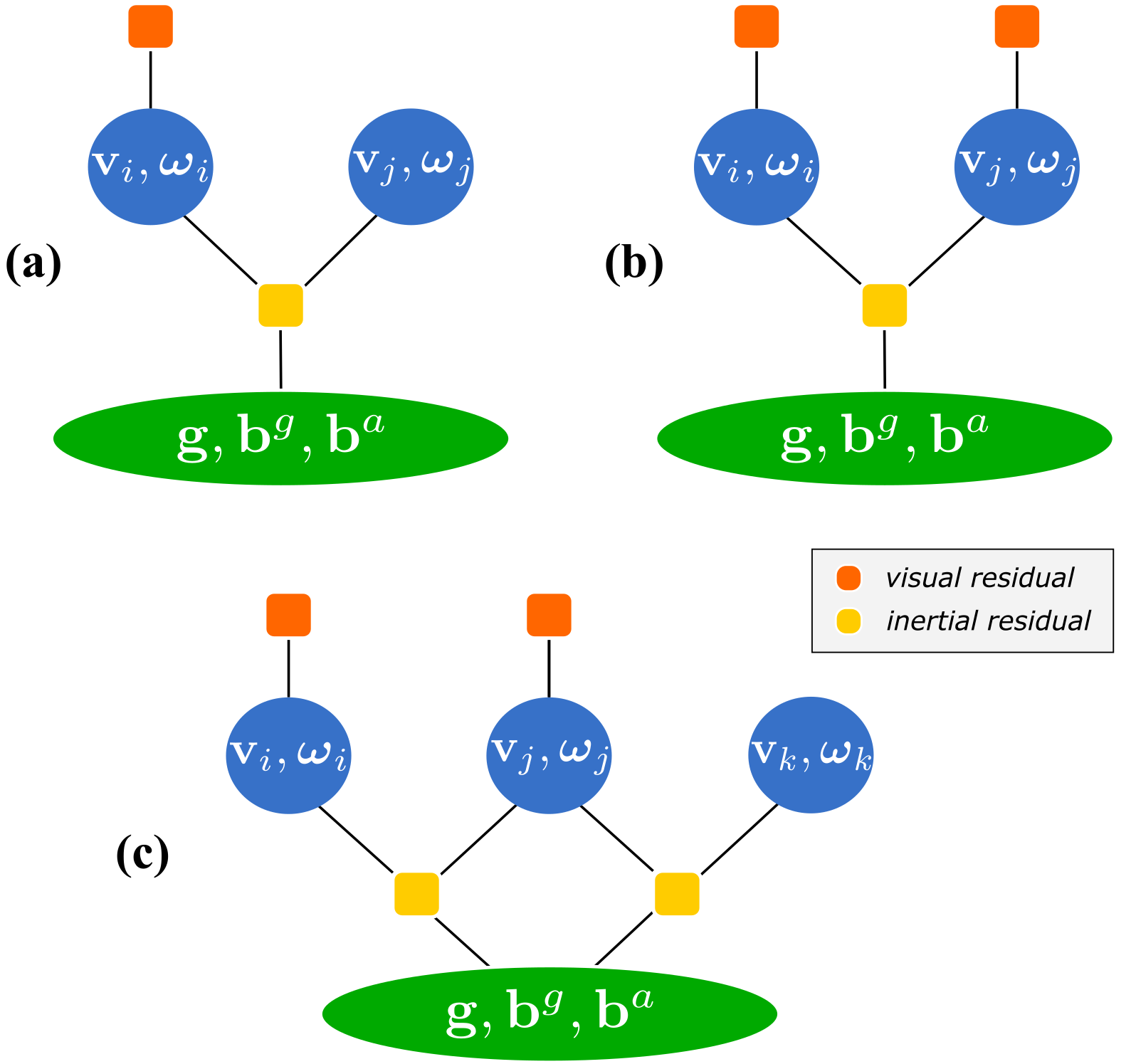}}
\caption{Factor graph representation using different modes of operation. Blue and green shapes contains the variables to be estimated. (a) Taking two frames, only one visual and one inertial residual are used. (b) Here we take three frames, so there are two visual residual. Only one inertial residual is used. (c) Is the same situation as before but in this case two inertial residual are used. The difference between (b) and (c) is the inertial constraint imposed by the last aggregate frame.}
\label{grafos-different-modes}
\end{figure}

Depending on the value of $N$, the system will have different \textit{operating modes} which will be detailed in this section. 
\newline \indent We start with the factor graph illustrated in Fig. \ref{grafos-different-modes}(a), where only frames $i$ and $j$ are available. Here, $N=2$ and any variable has been marginalized yet. 
Since we only have two frames, the cost function will be $J = J_i^{ij} + J_c^{ij}$. The first term is associated to the inertial measurements and the second one to the camera measurements.
Super-indices $^{ij}\cdot$ denote that the corresponding term is built up by frames $i$ and $j$. The state $\mathbf{x}$ contains in this case the velocities ${\mathbf{v}}_{i}, {\boldsymbol\omega}_{i}, {\mathbf{v}}_{j}$ and ${\boldsymbol\omega}_{j}$ as well as the gravity $^{W}\mathbf{g}$ and biases $\mathbf{b}^g$ and $\mathbf{b}^a$.

When a new frame  ($k-$frame) is added to the window, $N=3$ and the associated graph is shown in Fig. \ref{grafos-different-modes}(b). In this case the state remains the same as before but the cost changes to $J(\mathbf{x}) = J_i^{ij} + J_c^{ij} + J_{c}^{jk}$, \emph{i.e.}, only visual information is added to the cost function. 
\newline \indent The last case is when we add a new inertial term (between $j$ and $k$ frames), therefore $J(\mathbf{x}) = J_i^{ij} + J_c^{ij} + J_{c}^{jk} + J_i^{jk}$. This  situation is represented in Fig.  \ref{grafos-different-modes}(c). Here the state changes and the velocities ${\mathbf{v}}_{k}$ and $ {\boldsymbol\omega}_{k}$ are added.

In the general case, the state $\mathbf{x} \in \mathbb{R}^{6N+8}$ is defined as
\begin{equation}
        \mathbf{x} = \left[ {\mathbf{v}}_i^\top,{\boldsymbol\omega}_i^\top,\dots,{\mathbf{v}}_{i+N-1}^\top,{\boldsymbol\omega}_{i+N-1}^\top,{\mathbf{g}}^\top,{\mathbf{b}^g}^\top, {\mathbf{b}^a}^\top \right]^\top
\end{equation}
\noindent and the cost function $J(\mathbf{x})$ can be expressed compactly as follows
\begin{dmath}
        J(\mathbf{x})= 
             \sum_{p=i}^{i+N-1}\left(\mathbf{r}_{c_{p}}^\top{\boldsymbol\Sigma}_{c_{p}}^{-1}\mathbf{r}_{c_{p}} +  \mathbf{r}_{\Delta \mathbf{v}_{p}}^\top{\boldsymbol\Sigma}_{\Delta \mathbf{v}_{p}}^{-1} \mathbf{r}_{\Delta \mathbf{v}_{p}}\right)
            + \mathbf{r}_{bg}^\top{\boldsymbol\Sigma}_{\boldsymbol\omega}^{-1} \mathbf{r}_{bg}
            + \mathbf{r}_{ba}^\top{\boldsymbol\Sigma}_{\mathbf{a}}^{-1} \mathbf{r}_{ba}   
            + \sum_{l=i}^{i+N}\mathbf{r}_{\omega_l }^\top{\boldsymbol\Sigma}_{\boldsymbol\omega}^{-1}\mathbf{r}_{\omega_l}
\end{dmath}

\subsection{Marginalization}

\begin{figure}[h]
\centerline{\includegraphics[scale=1.83]{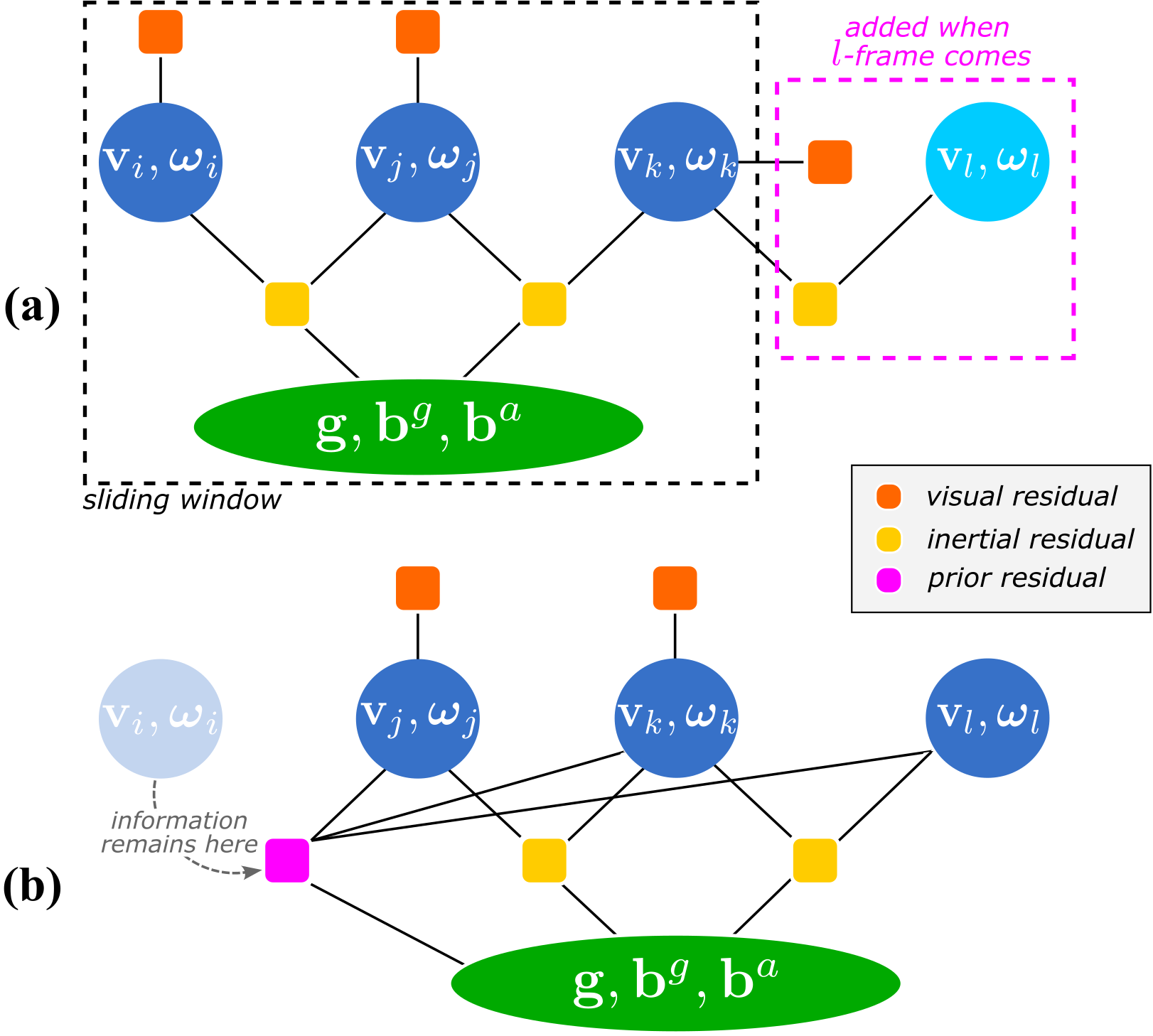}}
\caption{Factor graph using sliding window with containing 3 frames. Blue and green shapes contains the variables to be estimated. (a) When a new frame comes, both visual and inertial residual is added and the marginalization is done. (b) After marginalization, a new prior residual is added on the cost function.}
\label{grafo_marginalization}
\end{figure}

We mentioned that, as the sliding window moves, information from previous states is marginalized out.
Let consider the case in Fig. \ref{grafo_marginalization}, in which we want to perform 3-frames-sliding-window optimization.

Fig.~\ref{grafo_marginalization}(a) illustrates the graph after we have optimized the states at time steps $i$, $j$ and $k$, and a new frame ($l-$frame) arrives. The state estimate at this point, ${\mathbf{x}}_M$,  can be decomposed into two parts: the variables that we want to marginalize $\mathbf{x}_\alpha = [\mathbf{v}_i^\top, \boldsymbol{\omega}_i^\top]^\top\in \mathbb{R}^{6}$ and the starting point for the next step ${\mathbf{x}}_\beta = \left[{\mathbf{v}}_j^\top,{\boldsymbol\omega}_j^\top,{\mathbf{v}}_k^\top,{\boldsymbol\omega}_k^\top,{\mathbf{v}}_l^\top,{\boldsymbol\omega}_l^\top,{\mathbf{g}}^\top,{\mathbf{b}^g}^\top, {\mathbf{b}^a}^\top \right]^\top$. 
\begin{equation}
 {\mathbf{x}}_M
  =\begin{bMatrix}{c:c}
  \mathbf{x}_\alpha^\top & \mathbf{x}_\beta^\top
  \end{bMatrix}^\top
  \label{eq:r0-antes-de-marginalizacion}
\end{equation}


Consider Eq. (\ref{cost-function-marginalization}), where we have the cost function before marginalization $J(\mathbf{x})$, and a new term $J_{p}$, which stands for the marginalization priors and accounts for the information associated to the marginalized variables
\begin{equation}
        J^*(\mathbf{x})\doteq J(\mathbf{x}) + J_{p}
\label{cost-function-marginalization}
\end{equation}

Using second-order Taylor approximation, the cost $J(\mathbf{x})$ can be expressed as follows
\begin{equation}
        J(\mathbf{x}) \approx J(\mathbf{x}_0) 
        + \cancel{\nabla J(\mathbf{x}_0)\mathbf{r}}
        + {\frac{1}{2}\mathbf{r}^\top\mathbf{H}(\mathbf{x}_0)\mathbf{r}}
\label{eq:taylor-cost-function}        
\end{equation}

The above approximation is calculated around a state $\mathbf{x}_0$, where a minimum is achieved, \emph{i.e.} $\nabla J(\mathbf{x}_M)\mathbf{r}=0$ ($\mathbf{r} = \mathbf{x} - \mathbf{x}_0$).
The Hessian $\mathbf{H}$ 
contains the second derivatives of the cost function with respect to the state variables, therefore it 
encodes how every state variable affects the others. 

We denote as $\alpha$ the block of variables we would like to marginalize, and $\beta$ the block of variables we would like to keep.
When marginalizing a set $\alpha$ of variables, we gather all factors dependent on them as well as the connected variables $\beta$. This is done by means of the Schur Complement as follows 
\begin{equation}
        \mathbf{H}^* = \mathbf{H}_{\beta\beta} 
                     - \mathbf{H}_{\alpha\beta}^\top\mathbf{H}_{\alpha\alpha}^{-1}\mathbf{H}_{\alpha\beta}   
\label{eq:schur-complement}
\end{equation}

\begin{figure}[!h]
\centerline{\includegraphics[scale=1.13]{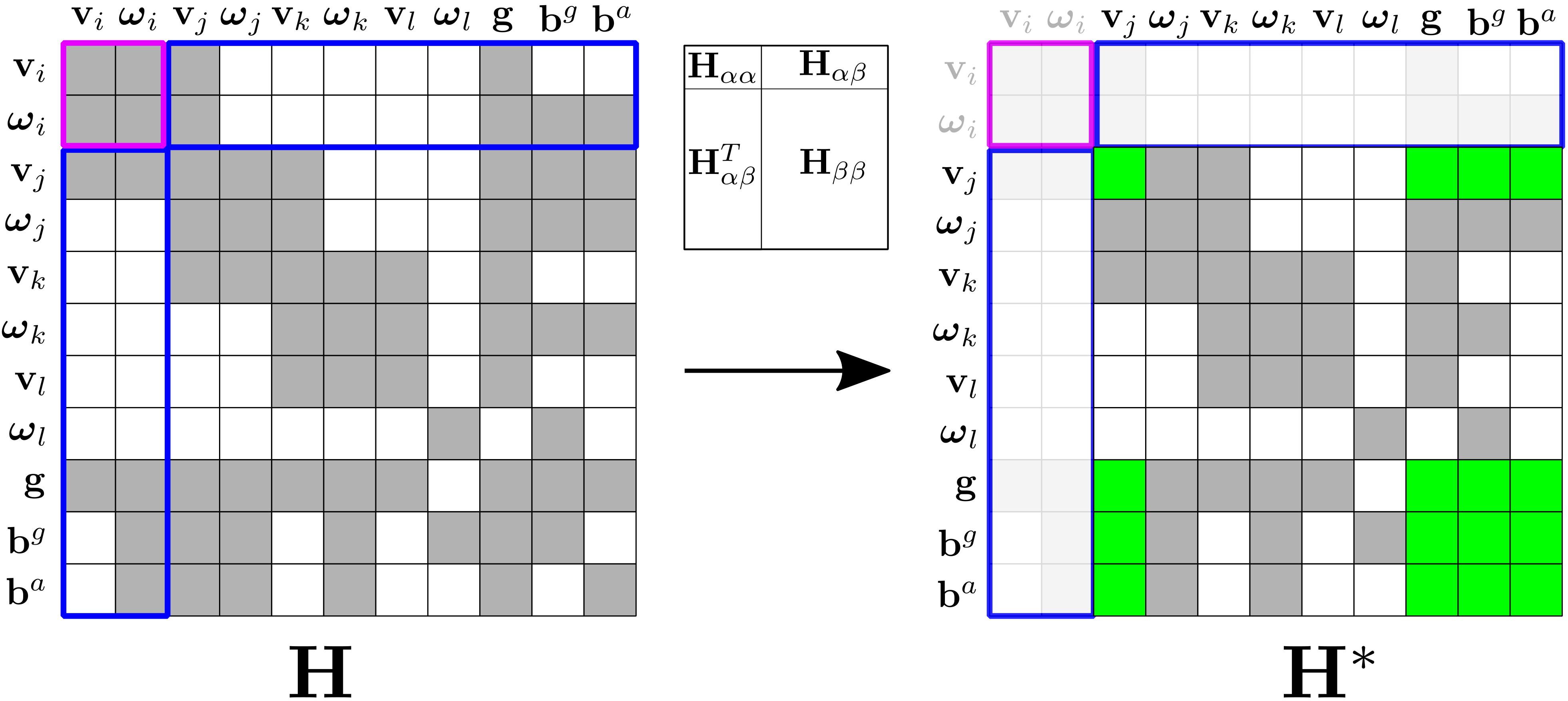}}
\caption{Marginalization example. We start with a Hessian matrix $\mathbf{H}$ after optimization with $N=4$. We want to marginalize $\mathbf{v}_i$ and $\boldsymbol{\omega}_i$. The marginalized Hessian matrix $\mathbf{H}^*$ corresponds to the Schur complement of $\mathbf{H}_{\alpha\alpha}$. This calculation transfers the information constraints of the variable being eliminated to its adjacent nodes, adding shared information between these variables (green cells).}
\label{optimization_marginalization}
\end{figure}

Fig. \ref{optimization_marginalization} graphically illustrates how the $\alpha-$block (variables $\mathbf{v}_i$ and $\boldsymbol{\omega}_{i}$) is removed from $\mathbf{H}$ but the information is preserved in the new matrix $\mathbf{H}^*$.

Considering ${\mathbf{x}}_\beta$ obtained from Eq. (\ref{eq:r0-antes-de-marginalizacion}) and the new state $\mathbf{x}^*$, the term $\mathbf{r}_{p}$ can be expressed as $\mathbf{r}_{p} = \mathbf{x}^* - {\mathbf{x}}_\beta$,
where
\begin{equation*}
        \mathbf{x}^* = \left[{\mathbf{v}}_j^\top,{\boldsymbol\omega}_j^\top,{\mathbf{v}}_k^\top,{\boldsymbol\omega}_k^\top,{\mathbf{v}}_l^\top,{\boldsymbol\omega}_l^\top,{\mathbf{g}}^\top,{\mathbf{b}^g}^\top, {\mathbf{b}^a}^\top \right]^\top 
\end{equation*}

\noindent The new term  $J_{p}$ can now be defined as follows
\begin{equation}
\label{eq:new-E-prior}
    J_{p} = \mathbf{r}_{p}^\top \hspace{0.3mm}{\mathbf{H}^*}\hspace{0.3mm}\mathbf{r}_{p}
\end{equation}

Finally, the result is obtained by minimizing the cost function $J^*(\mathbf{x})$ expressed in Eq. (\ref{cost-function-marginalization}).


\begin{table*}[!h]
\centering
\resizebox{17.4cm}{!} {
\begin{tabular}{|
>{\columncolor[HTML]{EFEFEF}}c ||
>{\columncolor[HTML]{FFFFFF}}c 
>{\columncolor[HTML]{FFFFFF}}c ||
>{\columncolor[HTML]{FFFFFF}}c 
>{\columncolor[HTML]{FFFFFF}}c 
>{\columncolor[HTML]{FFFFFF}}c ||
>{\columncolor[HTML]{FFFFFF}}c 
>{\columncolor[HTML]{FFFFFF}}c ||
>{\columncolor[HTML]{FFFFFF}}c 
>{\columncolor[HTML]{FFFFFF}}c 
>{\columncolor[HTML]{FFFFFF}}c ||}
\hline
\cellcolor[HTML]{FFFFFF}                                                       & \multicolumn{2}{c||}{\cellcolor[HTML]{FFFFFF}2-frames}                                                                                                                             & \multicolumn{3}{c||}{\cellcolor[HTML]{FFFFFF}3-frames}                                                                                                                                                                                                                                              & \multicolumn{2}{c||}{\cellcolor[HTML]{FFFFFF}4-frames}                                                                                                                             & \multicolumn{3}{c||}{\cellcolor[HTML]{FFFFFF}5-frames}                                                                                                                                                                                                                                              \\ \cline{2-11} 
\multirow{-2}{*}{\cellcolor[HTML]{FFFFFF}}                                     & \multicolumn{1}{c|}{\cellcolor[HTML]{EFEFEF}\begin{tabular}[c]{@{}c@{}}RGB-D \\ (DIFODO$^*$)\end{tabular}}                                                       & \cellcolor[HTML]{EFEFEF}\begin{tabular}[c]{@{}c@{}}RGB-D-I \\ (ours)\end{tabular}                                          & \multicolumn{1}{c|}{\cellcolor[HTML]{EFEFEF}\begin{tabular}[c]{@{}c@{}}RGB-D \\ ($^{**}$)\end{tabular}}                                                       & \multicolumn{1}{c|}{\cellcolor[HTML]{EFEFEF}\begin{tabular}[c]{@{}c@{}}RGB-D-I \\ (ours)\end{tabular} }                                                               & \cellcolor[HTML]{EFEFEF}\begin{tabular}[c]{@{}c@{}}RGB-D-I \\ (ours+M)\end{tabular}                                    & \multicolumn{1}{c|}{\cellcolor[HTML]{EFEFEF}\begin{tabular}[c]{@{}c@{}}RGB-D \\ ($^{**}$)\end{tabular}}                                                       & \cellcolor[HTML]{EFEFEF}\begin{tabular}[c]{@{}c@{}}RGB-D-I \\ (ours)\end{tabular}                                           & \multicolumn{1}{c|}{\cellcolor[HTML]{EFEFEF}\begin{tabular}[c]{@{}c@{}}RGB-D \\ ($^{**}$)\end{tabular}}                                                       & \multicolumn{1}{c|}{\cellcolor[HTML]{EFEFEF}\begin{tabular}[c]{@{}c@{}}RGB-D-I \\ (ours)\end{tabular}}                                                               & \cellcolor[HTML]{EFEFEF}\begin{tabular}[c]{@{}c@{}}RGB-D-I \\ (ours+M)\end{tabular}                                      \\ \hline
\begin{tabular}[c]{@{}c@{}}$\mathrm{RMSE}_{{\mathbf{v}}}$ \\ {[}$cm/s${]}\end{tabular}                    & \multicolumn{1}{c|}{\cellcolor[HTML]{FFFFFF}\begin{tabular}[c]{@{}c@{}}8.020\\ ±4.989\end{tabular}}     & \textbf{\begin{tabular}[c]{@{}c@{}}7.897\\ ±4.699\end{tabular}}     & \multicolumn{1}{c|}{\cellcolor[HTML]{FFFFFF}\begin{tabular}[c]{@{}c@{}}0.565\\ ±0.508\end{tabular}}     & \multicolumn{1}{c|}{\cellcolor[HTML]{FFFFFF}\begin{tabular}[c]{@{}c@{}}0.524\\ ±0.442\end{tabular}}          & \textbf{\begin{tabular}[c]{@{}c@{}}0.523\\ ±0.250\end{tabular}}     & \multicolumn{1}{c|}{\cellcolor[HTML]{FFFFFF}\begin{tabular}[c]{@{}c@{}}0.575\\ ±0.514\end{tabular}}     & \textbf{\begin{tabular}[c]{@{}c@{}}0.533\\ ±0.447\end{tabular}}     & \multicolumn{1}{c|}{\cellcolor[HTML]{FFFFFF}\begin{tabular}[c]{@{}c@{}}0.582\\ ±0.520\end{tabular}}     & \multicolumn{1}{c|}{\cellcolor[HTML]{FFFFFF}\begin{tabular}[c]{@{}c@{}}0.538\\ ±0.452\end{tabular}}          & \textbf{\begin{tabular}[c]{@{}c@{}}0.535\\ ±0.245\end{tabular}}     \\ \hline
\begin{tabular}[c]{@{}c@{}}$\mathrm{RMSE}_{{\boldsymbol\omega}}$\\ {[}$rad/s${]}\end{tabular}                   & \multicolumn{1}{c|}{\cellcolor[HTML]{FFFFFF}\begin{tabular}[c]{@{}c@{}}0.168\\ ±0.079\end{tabular}} & \textbf{\begin{tabular}[c]{@{}c@{}}0.037\\ ±0.022\end{tabular}} & \multicolumn{1}{c|}{\cellcolor[HTML]{FFFFFF}\begin{tabular}[c]{@{}c@{}}0.00113\\ ±0.00081\end{tabular}} & \multicolumn{1}{c|}{\cellcolor[HTML]{FFFFFF}\begin{tabular}[c]{@{}c@{}}0.00107\\ ±0.00072\end{tabular}}      & \textbf{\begin{tabular}[c]{@{}c@{}}0.00091\\ ±0.00052\end{tabular}} & \multicolumn{1}{c|}{\cellcolor[HTML]{FFFFFF}\begin{tabular}[c]{@{}c@{}}0.00114\\ ±0.00083\end{tabular}} & \textbf{\begin{tabular}[c]{@{}c@{}}0.00108\\ ±0.00074\end{tabular}} & \multicolumn{1}{c|}{\cellcolor[HTML]{FFFFFF}\begin{tabular}[c]{@{}c@{}}0.00116\\ ±0.00085\end{tabular}} & \multicolumn{1}{c|}{\cellcolor[HTML]{FFFFFF}\begin{tabular}[c]{@{}c@{}}0.00109\\ ±0.00075\end{tabular}}      & \textbf{\begin{tabular}[c]{@{}c@{}}0.00094\\ ±0.00054\end{tabular}} \\ \hline
\begin{tabular}[c]{@{}c@{}}$\mathrm{RMSE}_{\mathbf{b}^a}$\\ {[}$cm/s^2${]}\end{tabular} & \multicolumn{1}{c|}{\cellcolor[HTML]{FFFFFF}\textbf{-}}                                                   & \begin{tabular}[c]{@{}c@{}}0.177\\ ±0.026\end{tabular}              & \multicolumn{1}{c|}{\cellcolor[HTML]{FFFFFF}\textbf{-}}                                                   & \multicolumn{1}{c|}{\cellcolor[HTML]{FFFFFF}\textbf{\begin{tabular}[c]{@{}c@{}}0.214\\ ±0.083\end{tabular}}} & \begin{tabular}[c]{@{}c@{}}0.243\\ ±0.112\end{tabular}              & \multicolumn{1}{c|}{\cellcolor[HTML]{FFFFFF}\textbf{-}}                                                   & \begin{tabular}[c]{@{}c@{}}0.227\\ ±0.141\end{tabular}              & \multicolumn{1}{c|}{\cellcolor[HTML]{FFFFFF}\textbf{-}}                                                   & \multicolumn{1}{c|}{\cellcolor[HTML]{FFFFFF}\begin{tabular}[c]{@{}c@{}}0.217\\ ±0.171\end{tabular}}          & \textbf{\begin{tabular}[c]{@{}c@{}}0.199\\ ±0.080\end{tabular}}     \\ \hline
\begin{tabular}[c]{@{}c@{}}$\mathrm{RMSE}_{\mathbf{b}^g}$\\ {[}$rad/s${]}\end{tabular}                  & \multicolumn{1}{c|}{\cellcolor[HTML]{FFFFFF}\textbf{-}}                                                   & \begin{tabular}[c]{@{}c@{}}0.031\\ ±0.025\end{tabular}              & \multicolumn{1}{c|}{\cellcolor[HTML]{FFFFFF}\textbf{-}}                                                   & \multicolumn{1}{c|}{\cellcolor[HTML]{FFFFFF}\textbf{\begin{tabular}[c]{@{}c@{}}0.048\\ ±0.100\end{tabular}}} & \begin{tabular}[c]{@{}c@{}}0.049\\ ±0.012\end{tabular}              & \multicolumn{1}{c|}{\cellcolor[HTML]{FFFFFF}\textbf{-}}                                                   & \begin{tabular}[c]{@{}c@{}}0.086\\ ±0.168\end{tabular}              & \multicolumn{1}{c|}{\cellcolor[HTML]{FFFFFF}\textbf{-}}                                                   & \multicolumn{1}{c|}{\cellcolor[HTML]{FFFFFF}\textbf{\begin{tabular}[c]{@{}c@{}}0.098\\ ±0.209\end{tabular}}} & \begin{tabular}[c]{@{}c@{}}0.103\\ ±0.073\end{tabular}              \\ \hline
\begin{tabular}[c]{@{}c@{}}$\theta_{\textbf{g}}$\\ {[}$rad${]}\end{tabular}                     & \multicolumn{1}{c|}{\cellcolor[HTML]{FFFFFF}\textbf{-}}                                                   & \begin{tabular}[c]{@{}c@{}}0.372\\ ±0.308\end{tabular}              & \multicolumn{1}{c|}{\cellcolor[HTML]{FFFFFF}-}                                                            & \multicolumn{1}{c|}{\cellcolor[HTML]{FFFFFF}\begin{tabular}[c]{@{}c@{}}0.299\\ ±0.236\end{tabular}}          & \textbf{\begin{tabular}[c]{@{}c@{}}0.168\\ ±0.089\end{tabular}}     & \multicolumn{1}{c|}{\cellcolor[HTML]{FFFFFF}-}                                                            & \begin{tabular}[c]{@{}c@{}}0.273\\ ±0.204\end{tabular}              & \multicolumn{1}{c|}{\cellcolor[HTML]{FFFFFF}-}                                                            & \multicolumn{1}{c|}{\cellcolor[HTML]{FFFFFF}\begin{tabular}[c]{@{}c@{}}0.275\\ ±0.215\end{tabular}}          & \textbf{\begin{tabular}[c]{@{}c@{}}0.167\\ ±0.086\end{tabular}}     \\ \hline
\end{tabular}
}
\caption{Error Metrics on ICL- NUIM for different operating modes. (DIFODO$^*$) stands for our implementation of the method in \cite{jaimez2015fast}. ($^{**}$) indicates the output of our DIFODO$^*$ implementation between pairs of frames.}
\label{Table-performance-different-modes}
\end{table*}
\begin{figure*}[]
\centerline{\includegraphics[scale=1.95]{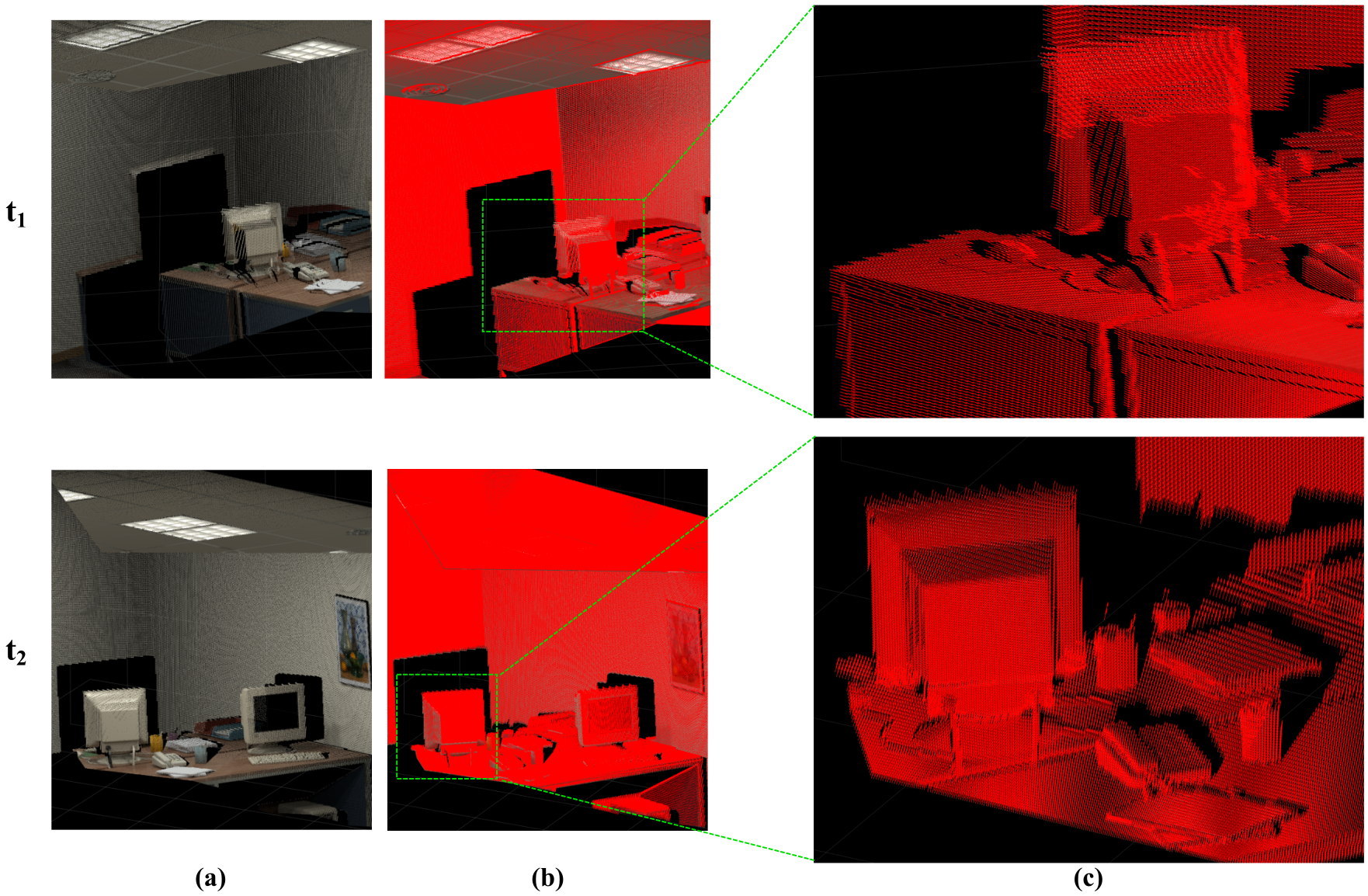}}
\caption{Motion estimation in an office scene from the ICL-NUIM dataset in two different times $t_1$ and $t_2$. (a) 3D representation of the scene. (b) Motion estimation of the objects in the scene. Every velocity is represented by a red arrow on each point. (c) Zoomed-in areas.}
\label{motion-representation}
\end{figure*}
\begin{table*}[!h]
\centering
\resizebox{13.5cm}{!} {

\begin{tabular}{|
>{\columncolor[HTML]{EFEFEF}}c ||
>{\columncolor[HTML]{FFFFFF}}c 
>{\columncolor[HTML]{FFFFFF}}c ||
>{\columncolor[HTML]{FFFFFF}}c 
>{\columncolor[HTML]{FFFFFF}}c ||
>{\columncolor[HTML]{FFFFFF}}c 
>{\columncolor[HTML]{FFFFFF}}c ||
>{\columncolor[HTML]{FFFFFF}}c 
>{\columncolor[HTML]{FFFFFF}}c ||}
\hline
\cellcolor[HTML]{FFFFFF}                                                        & \multicolumn{2}{c||}{\cellcolor[HTML]{FFFFFF}2-frames}                                                                                                                                   & \multicolumn{2}{c||}{\cellcolor[HTML]{FFFFFF}3-frames}                                                                                                                             & \multicolumn{2}{c||}{\cellcolor[HTML]{FFFFFF}4-frames}                                                                                                                             & \multicolumn{2}{c||}{\cellcolor[HTML]{FFFFFF}5-frames}                                                                                                                             \\ \cline{2-9} 
\multirow{-2}{*}{\cellcolor[HTML]{FFFFFF}}                                      & \multicolumn{1}{c|}{\cellcolor[HTML]{EFEFEF}\begin{tabular}[c]{@{}c@{}}RGB-D \\ ($^{**}$)\end{tabular}}                                                                & \cellcolor[HTML]{EFEFEF}\begin{tabular}[c]{@{}c@{}}RGB-D-I \\ (ours)\end{tabular}                                        & \multicolumn{1}{c|}{\cellcolor[HTML]{EFEFEF}\begin{tabular}[c]{@{}c@{}}RGB-D \\ (\textit{DIFODO}$^*$)\end{tabular}}                                                       & \cellcolor[HTML]{EFEFEF}\begin{tabular}[c]{@{}c@{}}RGB-D-I \\ (ours)\end{tabular}                                           & \multicolumn{1}{c|}{\cellcolor[HTML]{EFEFEF}\begin{tabular}[c]{@{}c@{}}RGB-D \\ ($^{**}$)\end{tabular}}                                                       & \cellcolor[HTML]{EFEFEF}\begin{tabular}[c]{@{}c@{}}RGB-D-I \\ (ours)\end{tabular}                                          & \multicolumn{1}{c|}{\cellcolor[HTML]{EFEFEF}\begin{tabular}[c]{@{}c@{}}RGB-D \\ ($^{**}$)\end{tabular}}                                                       & \cellcolor[HTML]{EFEFEF}\begin{tabular}[c]{@{}c@{}}RGB-D-I \\ (ours)\end{tabular}                                          \\ \hline
\begin{tabular}[c]{@{}c@{}}$\mathrm{RMSE}_{{\mathbf{v}}}$ \\ {[}$cm/s${]}\end{tabular}                    & \multicolumn{1}{c|}{\cellcolor[HTML]{FFFFFF}\begin{tabular}[c]{@{}c@{}}36.993\\ ±11.295\end{tabular}}            & \textbf{\begin{tabular}[c]{@{}c@{}}34.008\\ ±8.438\end{tabular}} & \multicolumn{1}{c|}{\cellcolor[HTML]{FFFFFF}\begin{tabular}[c]{@{}c@{}}22.453\\ ±9.029\end{tabular}}    & \textbf{\begin{tabular}[c]{@{}c@{}}22.054\\ ±8.766\end{tabular}}    & \multicolumn{1}{c|}{\cellcolor[HTML]{FFFFFF}\begin{tabular}[c]{@{}c@{}}22.453\\ ±9.029\end{tabular}}    & \textbf{\begin{tabular}[c]{@{}c@{}}21.808\\ ±8.646\end{tabular}}    & \multicolumn{1}{c|}{\cellcolor[HTML]{FFFFFF}\begin{tabular}[c]{@{}c@{}}22.453\\ ±9.029\end{tabular}}    & \textbf{\begin{tabular}[c]{@{}c@{}}21.924\\ ±8.769\end{tabular}}    \\ \hline
\begin{tabular}[c]{@{}c@{}}$\mathrm{RMSE}_{{\boldsymbol\omega}}$\\ {[}$rad/s${]}\end{tabular}                    & \multicolumn{1}{c|}{\cellcolor[HTML]{FFFFFF}\textbf{\begin{tabular}[c]{@{}c@{}}0.165\\ ±0.090\end{tabular}}} & \begin{tabular}[c]{@{}c@{}}0.184\\ ±0.102\end{tabular}       & \multicolumn{1}{c|}{\cellcolor[HTML]{FFFFFF}\begin{tabular}[c]{@{}c@{}}0.172\\ ±0.091\end{tabular}} & \textbf{\begin{tabular}[c]{@{}c@{}}0.169\\ ±0.092\end{tabular}} & \multicolumn{1}{c|}{\cellcolor[HTML]{FFFFFF}\begin{tabular}[c]{@{}c@{}}0.172\\ ±0.091\end{tabular}} & \textbf{\begin{tabular}[c]{@{}c@{}}0.168\\ ±0.092\end{tabular}} & \multicolumn{1}{c|}{\cellcolor[HTML]{FFFFFF}\begin{tabular}[c]{@{}c@{}}0.172\\ ±0.091\end{tabular}} & \textbf{\begin{tabular}[c]{@{}c@{}}0.167\\ ±0.092\end{tabular}} \\ \hline
\begin{tabular}[c]{@{}c@{}}$\mathrm{RMSE}_{\mathbf{b}^a}$\\ {[}$cm/s^2${]}\end{tabular} & \multicolumn{1}{c|}{\cellcolor[HTML]{FFFFFF}-}                                                                     & \begin{tabular}[c]{@{}c@{}}0.121\\ ±0.001\end{tabular}           & \multicolumn{1}{c|}{\cellcolor[HTML]{FFFFFF}-}                                                            & \begin{tabular}[c]{@{}c@{}}0.102\\ ±0.028\end{tabular}              & \multicolumn{1}{c|}{\cellcolor[HTML]{FFFFFF}-}                                                            & \begin{tabular}[c]{@{}c@{}}0.097\\ ±0.014\end{tabular}              & \multicolumn{1}{c|}{\cellcolor[HTML]{FFFFFF}-}                                                            & \begin{tabular}[c]{@{}c@{}}0.113\\ ±0.024\end{tabular}              \\ \hline
\begin{tabular}[c]{@{}c@{}}$\mathrm{RMSE}_{\mathbf{b}^g}$\\ {[}$rad/s${]}\end{tabular}                   & \multicolumn{1}{c|}{\cellcolor[HTML]{FFFFFF}-}                                                                     & \begin{tabular}[c]{@{}c@{}}0.015\\ ±0.001\end{tabular}           & \multicolumn{1}{c|}{\cellcolor[HTML]{FFFFFF}-}                                                            & \begin{tabular}[c]{@{}c@{}}0.015\\ ±0.001\end{tabular}              & \multicolumn{1}{c|}{\cellcolor[HTML]{FFFFFF}-}                                                            & \begin{tabular}[c]{@{}c@{}}0.015\\ ±0.001\end{tabular}              & \multicolumn{1}{c|}{\cellcolor[HTML]{FFFFFF}-}                                                            & \begin{tabular}[c]{@{}c@{}}0.015\\ ±0.001\end{tabular}              \\ \hline
\begin{tabular}[c]{@{}c@{}}$\theta_{\textbf{g}}$\\ {[}$rad${]}\end{tabular}                      & \multicolumn{1}{c|}{\cellcolor[HTML]{FFFFFF}-}                                                                     & \begin{tabular}[c]{@{}c@{}}1.550\\ ±0.440\end{tabular}           & \multicolumn{1}{c|}{\cellcolor[HTML]{FFFFFF}-}                                                            & \begin{tabular}[c]{@{}c@{}}0.150\\ ±0.104\end{tabular}              & \multicolumn{1}{c|}{\cellcolor[HTML]{FFFFFF}-}                                                            & \begin{tabular}[c]{@{}c@{}}0.115\\ ±0.072\end{tabular}              & \multicolumn{1}{c|}{\cellcolor[HTML]{FFFFFF}-}                                                            & \begin{tabular}[c]{@{}c@{}}0.095\\ ±0.061\end{tabular}              \\ \hline
\end{tabular}
}
\caption{Error Metrics on OpenLORIS-Scene for different operating modes. (DIFODO$^*$) stands for our implementation of the method in \cite{jaimez2015fast}. ($^{**}$) indicates the output of our DIFODO$^*$ implementation between pairs of frames.}
\label{Table-performance-Openloris-different-modes}

\end{table*}

\section{Experiments}


\subsection{Setup}

Public benchmarks that provide IMU, color and depth images are scarce. We chose to evaluate our proposal on an extended version of the living room sequences in the ICL-NUIM dataset~\cite{ICL-NUIM-dataset}. ICL-NUIM is a synthetic photorealistic dataset that provides ground truth poses as well as 3D scene models to benchmark reconstruction and/or localization approaches. As ICL-NUIM does not provide IMU data, in a manner similar to \cite{dense-kerl}, we fit splines to the ground truth poses to simulate continuous trajectories and simulated IMU measurements from them. We use the IMU model described in \cite{max-nikolic} and the same IMU parameters as \cite{dense-laidlow}.
We also evaluated our RGB-D-inertial flow in the OpenLORIS-Scene datasets \cite{openloris_dataset}, in which data are collected in real-world indoor scenes, for multiple times in each place to include natural scene changes in everyday scenarios. RGB-D images and IMU measurements from a RealSense D435i are provided. The ground truth trajectory was recorded by an OptiTrack MCS, that tracked artificial markers deployed on the Segway robot used to record the data. 

As metrics, we use the Root-Mean-Square-Error ($\mathrm{RMSE}$) for the velocities $\mathbf{v}$ and ${\boldsymbol\omega}$ and biases $\mathbf{b}^a$ and $\mathbf{b}^g$. 
For the gravity vector, we use the angle $\theta_{\textbf{g}} = \mathrm{cos}^{-1}\left(\sfrac{\hat{\textbf{g}} \cdot \textbf{g}_{gt}}{\|\hat{\textbf{g}}\|\|\textbf{g}_{gt}\|}\right)$ between the ground truth and estimated gravity direction. 



\subsection{Results}

\textbf{ICL-NUIM. }In this experiment we use the living room sequences in the ICL-NUIM dataset, and we run our RGB-D-I flow based method against the so-called DIFODO \cite{jaimez2015fast}, based on RGB-D flow. Note that, as \cite{jaimez2015fast} does not provide code, we used our own implementation based on the description in the paper. We run both scene flow methods for different estimation modes: 2-frames, 3-frames, 4-frames and 5-frames. Marginalization is not done here. The results are obtained over the entire dataset, taking as starting frame one out of every 2 which gives us more than 400 subsequences. In this experiment we have made 10 runs on the complete dataset.  
The results (specifically, the mean ± the standard deviation of the RMSE for the estimated states) are shown in Table \ref{Table-performance-different-modes}, in which the best result per estimation mode is boldfaced. Note that using inertial measurements improves the accuracy in both linear and angular velocity estimates. We can also observe that the errors are higher for the 2-frames case than for the rest, which shows how the information from additional frames is leveraged to estimate the inertial states. Note also how the standard deviation of the errors is reduced when inertial sensing is used, which indicates a higher robustness in challenging cases. 

\textbf{ICL-NUIM + Marginalization. }In this experiment, we show the effect of the marginalization in the living room sequences of ICL-NUIM, using two different sliding windows: 3 frames and 5 frames. In both cases we marginalize one frame only. 
The results are shown in the columns of Table \ref{Table-performance-different-modes} titled as \textit{RGB-D-I (ours+M)}. It can be seen how the gravity vector estimation is improved in comparison to the case without marginalization. 
Also, a small improvement occurs in the linear and angular velocities errors. 

\indent Fig. \ref{motion-representation} shows qualitative results for motion estimation. The office scene consists  on a computer on a desk and luminaires. Fig. \ref{motion-representation}(a) shows a point cloud extracted from the RGB-D data, and Fig. \ref{motion-representation}(b) displays the scene flow (red arrows represent the velocity in each point). 
For better appreciation, Fig. \ref{motion-representation}(c) zooms in some areas. Observe that our approach estimates a smooth flow even in textureless areas such as the background wall.

\textbf{OpenLORIS-Scene. }In this experiment we consider the office-1 sequence in the OpenLORIS-Scene Dataset, where the robot moves along a U-shape route.
As in previous experiments, we compare our RGB-D-I flow-based motion estimation against RGB-D-only motion estimation in 4 different optimization modes: 2-frames, 3-frames, 4-frames and 5-frames, all of them without marginalization.
The experiment is performed over the entire dataset, using one frame out of every five as starting point, which gave us more than 400 experiments. 
Table \ref{Table-performance-Openloris-different-modes} shows the mean ± the standard deviation of the RMSE for the results of such experiments. As in the synthetic case, it can be observed that using
inertial measurements improves the estimation results in both linear and angular velocities. It can also be observed how, as the optimization window grows, the errors of the inertial states are also smaller.

\section{Conclusions} 
\label{sec:conclusion}

In this work we present a novel camera motion estimation based on RGB-D-I scene flow. Specifically, we formulate the fusion of RGB-D and inertial data as a joint optimization using scene flow residuals and pre-integrated IMU residuals, weighted by their corresponding covariances. We also consider the marginalization of old states in order to keep a compact optimization. 
We evaluated our approach on a synthetic dataset, ICL-NUIM, and on a real dataset, OpenLORIS, both publicly available. Our results quantify the improvement that inertial fusion can offer to RGB-D scene flow techniques. 


{
    \small
    \bibliographystyle{ieeenat_fullname}

}


\end{document}